\documentclass[10pt,twocolumn,letterpaper]{article}

\usepackage{cvpr}      
\usepackage{colortbl}
\usepackage{fancyhdr}
\usepackage{amssymb}
\usepackage{bbm}
\usepackage{pifont}
\newcommand{\cmark}{\ding{51}}%
\newcommand{\xmark}{\textcolor{gray}{\ding{55}}}   
\usepackage{tablefootnote}
\usepackage{threeparttable}
\usepackage[accsupp]{axessibility}
\usepackage{tikz} 

\newcommand*\blackcircle{\raisebox{0.1ex}{\tikz{\node[shape=circle, draw=black, fill=black, inner sep=2pt] {};}}}
\newcommand*\whitecircle{\raisebox{0.1ex}{\tikz{\node[shape=circle, draw=black, fill=white, inner sep=2pt] {};}}}

\usepackage{xcolor}

\definecolor{darkgreen}{RGB}{34, 139, 34}

\newcolumntype{L}[1]{>{\raggedright\arraybackslash}p{#1}}

\definecolor{cvprblue}{rgb}{0.21,0.49,0.74}
\usepackage[pagebackref,breaklinks,colorlinks,allcolors=cvprblue]{hyperref}

\def\paperID{61} 
\def\confName{CVPR}
\def\confYear{2025}

\title{Action Valuation in Sports: A Survey}

\author{Artur Xarles$^{1,2}$\hspace{0.7cm} Sergio Escalera$^{1,2,3}$\hspace{0.7cm} Thomas B. Moeslund$^{3}$\hspace{0.7cm} Albert Clapés$^{1,2}$ \\
$^{1}$Universitat de Barcelona, Barcelona, Spain\\
$^{2}$Computer Vision Center, Cerdanyola del Vallès, Spain \\
$^{3}$Aalborg University, Aalborg, Denmark\\
{\tt\small arturxe@gmail.com}, {\tt\small sescalera@ub.edu}, {\tt\small tbm@create.aau.dk}, {\tt\small aclapes@ub.edu} \\
}

\begin{document}
\maketitle
\begin{abstract}
Action Valuation (AV) has emerged as a key topic in Sports Analytics, offering valuable insights by assigning scores to individual actions based on their contribution to desired outcomes. Despite a few surveys addressing related concepts such as Player Valuation, there is no comprehensive review dedicated to an in-depth analysis of AV across different sports. In this survey, we introduce a taxonomy with nine dimensions related to the AV task, encompassing data, methodological approaches, evaluation techniques, and practical applications. Through this analysis, we aim to identify the essential characteristics of effective AV methods, highlight existing gaps in research, and propose future directions for advancing the field.
\end{abstract}

\vspace{-0.6cm}
\section{Introduction}
\label{sec:intro}


While basic statistics have long been used in sports, the rise of sports analytics began in the early 2000s, driven by the Moneyball~\cite{lewis2004moneyball} phenomenon in baseball. This data-driven approach rapidly spread across sports, shaping today’s landscape where professional sports are intrinsically linked to advanced data analytics. The growing demand for data has driven the development of numerous computer vision tasks~\cite{thomas2017computer, naik2022comprehensive}, enabling the extraction of valuable insights such as player and ball tracking, as well as action localizations. Beyond data extraction, these developments support a wide range of analytical applications~\cite{ghosh2023sports}, including player and team performance evaluation\cite{carling2008performance}, injury prevention~\cite{schiff2010injury}, game strategy optimization, and referee assistance~\cite{held2023vars}.

One emerging challenge in sports analytics is \textbf{Action Valuation} (AV), illustrated in Figure~\ref{fig:taskAV}. AV aims to assign scores to individual actions based on their contribution to desired outcomes—such as a key pass increasing scoring chances or a last-man tackle preventing a goal in football. Unlike traditional performance metrics, AV offers a more fine-grained, context-aware assessment, making it valuable for evaluating team dominance, assessing player performance for scouting, and aiding decision-making through alternative action recommendations. However, the field faces two key difficulties: the scarcity of extensive, publicly available datasets (see Section~\ref{sec:data}), and the lack of a standardized evaluation framework due to the absence of explicit ground-truth annotations (see Section~\ref{sec:evaluation}). These limitations hinder the comparability of methods and the establishment of benchmarks, making it difficult to assess desirable characteristics of AV approaches and slowing progress in both research and commercial applications. 

Despite its growing relevance, AV remains an area with many open research questions. Existing surveys focus on related topics like Player Valuation~\cite{paper29, paper35, paper41, paper47}, often mentioning AV only briefly. They are also sport-specific, covering football~\cite{paper35, paper41, paper47} or basketball~\cite{paper29}, without offering a comprehensive analysis across sports. While some studies attempt to compare valuation approaches~\cite{paper4}, there is no dedicated survey that systematically examines AV methodologies. Thus, this paper aims to fill this gap by providing a comprehensive survey of Action Valuation in sports. Throughout, we use terms such as player, ball, and field, common in major sports, while noting that they may not apply to all sports. Our contributions include:

\begin{figure*}
    \centering
    \includegraphics[width=1\linewidth]{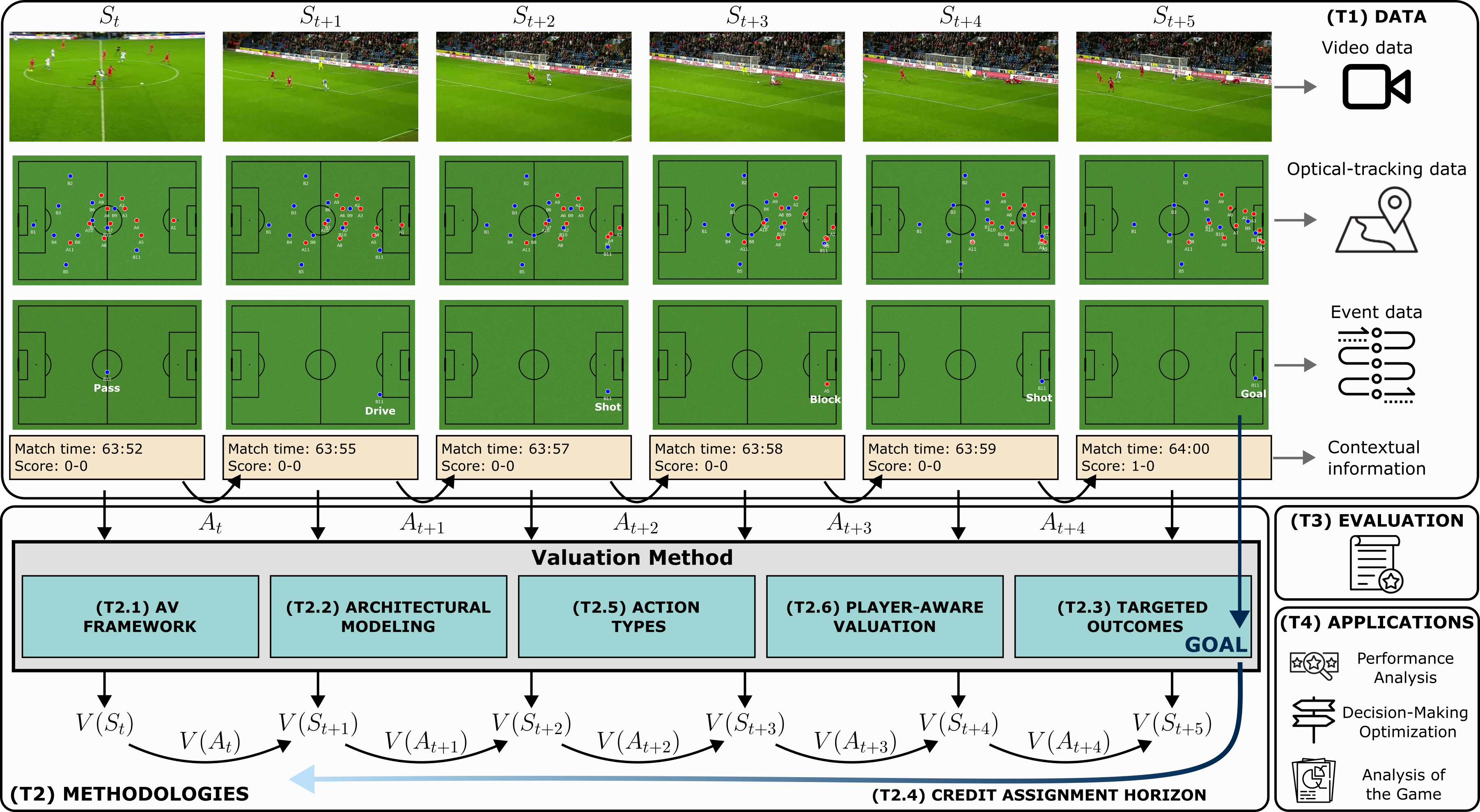}
    \caption{Illustration of the Action Valuation task, highlighting the nine dimensions of the taxonomy studied in this paper: (T1) Data, (T2) Methodological aspects, including (T2.1) AV Framework, (T2.2) Architectural Modeling, (T2.3) Targeted Outcomes, (T2.4) Credit Assignment Horizon, (T2.5) Action Types, and (T2.6) Player-Aware Valuation, (T3) Evaluation, and (T4) Applications. $S$, $A$, and $V(x)$ denote states, actions, and value of $x$, respectively.}
    \label{fig:taskAV}
    \vspace{-0.5cm}
\end{figure*}

\begin{enumerate}
\item A detailed overview of the datasets used for AV tasks, covering both publicly available and private datasets. While much of the existing work has focused on sports such as football, basketball, and ice hockey, we expand the scope by including a more diverse range of sports.
\item A systematical review and categorization of the most influential AV methodologies, structured by the proposed taxonomy with nine primary dimensions: (T1) Data, (T2) Methodological Aspects, which includes: (T2.1) AV Framework, (T2.2) Architectural Modeling, (T2.3) Targeted Outcomes, (T2.4) Credit Assignment Horizon (CAH), (T2.5) Action Types, and (T2.6) Player-Aware Valuation, (T3) Evaluation, and (T4) Applications. This taxonomy offers a structured framework for understanding the diverse approaches and their interrelationships.

\item We thoroughly analyze the proposed taxonomy and provide recommendations on the key characteristics AV methods should exhibit to produce meaningful valuations. Additionally, we discuss the main challenges in the AV task, such as the sparsity of desired outcomes in some sports, and the valuation of off-ball actions.

\item We discuss current gaps in AV research, such as the scarcity of comprehensive public datasets and the lack of standardized evaluation metrics, hindering AV research progress. By analyzing these gaps, we aim to identify promising directions for future work. 
\end{enumerate}

\noindent In the following sections, we define the Action Valuation task (Section~\ref{sec:task}), analyze the data (Section~\ref{sec:data}), discuss the taxonomy's methodological dimensions (Section~\ref{sec:methodologies}), review evaluation methods (Section~\ref{sec:evaluation}), explore task applications (Section~\ref{sec:applications}), and provide a general discussion (Section~\ref{sec:discussion}).







\section{Task Definition}
\label{sec:task}


\textbf{Action Valuation (AV)} in sports, illustrated in Figure~\ref{fig:taskAV}, involves assigning scores to individual actions performed by players based on their contribution to desirable outcomes. These outcomes can include scoring, maintaining possession of the ball, reaching key areas of the field, or increasing win probability. More formally, in Action Valuation, temporal sequences (such as games) can be divided into discrete states $S_t$, which represent the state of the game (e.g., positions of players, ball location, current score, etc.) at timestep $t$. The temporal difference between states can be defined by the time between consecutive actions, fixed frame steps when recording game data, or other approaches.


Transitions from one state to another occur due to a set of actions performed by players on the field, represented as $S_t \xrightarrow{A_t} S_{t+1}$, where $A_t = \{a_t^1, a_t^2, a_t^3, \dots \}$ denotes the set of actions taken between $S_t$ and $S_{t+1}$. These actions can include off-ball actions, such as player movements, or on-ball actions, such as passes or shots. The objective is to quantify the value of these actions. While some methods directly compute this value as $V(A_t) = V(A_t\mid S_t)$, others introduce an intermediate process of \textbf{State Valuation (SV)}, where each state is assigned a value $V(S_t)$ reflecting the likelihood of achieving desirable outcomes in future states. This is usually referred to as Expected Possession Value (EPV) when considering future states within the ongoing possession. In this approach, action values are derived from state values ---for example, as the change in value between consecutive states, $V(A_t) = V(S_{t+1}) - V(S{t})$. For both approaches, the distribution of $V(A_t)$ across individual actions $\{a_t^1, a_t^2, a_t^3, \dots \}$, as well as the method of assigning action or state values, depend on the approach used.

\vspace{-0.4cm}
\paragraph{Related Tasks.} Action Valuation closely relates to \textit{Action Quality Assessment (AQA)}~\cite{pirsiavash2014assessing}, as both aim to assign scores to actions. However, AQA evaluates actions based on human-annotated scores as ground truth. In contrast, Action Valuation relies on future outcomes to derive scores, operating without explicit ground-truth annotations. Another related task is \textit{Player Valuation (PV)}~\cite{terner2021modeling}, which quantifies a player's value rather than individual actions. As in AV, PV can be based on performance, but it may also consider market value~\cite{franceschi2024determinants}, influenced by factors such as age and popularity. Additionally, tasks like \textit{Game State Reconstruction}~\cite{somers2024soccernet}, which integrates player tracking, reidentification, and pitch location, can assist AV by providing high-level game state representations. Similarly, \textit{Action Spotting}~\cite{giancola2024deep} can help in localizing actions that drive state transitions.



\section{Data}
\label{sec:data}

Data for AV can be classified into three types (see Figure~\ref{fig:taskAV}):

\vspace{-0.4cm}
\paragraph{Video data (VD).} Video data offers a comprehensive view of the field, capturing the perception of player positions and poses. However, it can miss information occluded or off-screen information. Its high dimensionality and low-level nature can also complicate the learning process. 

\vspace{-0.4cm}
\paragraph{Optical-tracking data (OTD).} OTD tracks player and ball movements, mapping them onto a 2D representation of the field. It offers higher-level information that facilitates model training and allows for the automatic detection of off-ball actions, such as player movements. However, it lacks key details such as players appearances and specific poses, which could help identify individual players.

\vspace{-0.4cm}
\paragraph{Event data (ED).} ED includes annotations for on-ball actions, specifying their type, timing, location, and the player involved. While essential for AV, it lacks broader context like player positions and off-ball actions. However, additional \textbf{contextual information}, such as game score, period, or minutes played, can be directly derived from event data.


Given the data types and task requirements, an ideal dataset for AV would combine ED for on-ball actions with OTD for comprehensive game state representation and off-ball action detection. Extracting player poses from VD and additional contextual information from ED would also be beneficial. Furthermore, the dataset should include consistent player identification across multiple games, essential for specific applications (see Section~\ref{sec:applications}) and evaluation methods (see Section~\ref{sec:evaluation}), a sufficient number of games to cover diverse scenarios and duels of varying skill levels, and  temporal consistency to track the evolution of action, player, and team scores over time.

\begin{table}[t]
  \centering
  \rowcolors{4}{gray!25}{white}
  \begin{threeparttable}
  \resizebox{\columnwidth}{!}{
  \begin{tabular}{lccccccccc}
    \toprule
    Dataset & Sport & Public & \multicolumn{3}{c}{Data Type} & Games \\
    \cline{4-6}
    \multicolumn{3}{c}{} & VD & OTD & ED \\
    \bottomrule
    
    \textcolor{darkgreen}{[D1]} - StatsBomb\textsuperscript{\dag}\cite{StatsBombEvent} & Football & \cmark & \xmark & \xmark & \cmark & 3433 \\
    
    \textcolor{darkgreen}{[D2]} - StatsBomb 360 \cite{StatsBomb360} & Football & \cmark & \xmark & \cmark\tnote{1} & \cmark & 394 \\
    
    \textcolor{darkgreen}{[D3]} - Belgian Pro League~\cite{rahimian2022beyond} & Football & \xmark & \xmark & \cmark & \cmark & 430 \\
    
    \textcolor{darkgreen}{[D4]} - Meiji J1 League~\cite{paper12} & Football & \xmark & \xmark & \cmark & \cmark & 55 \\
    
    \textcolor{darkgreen}{[D5]} - STATS LLC~\cite{paper15} & Football & \xmark & \xmark & \cmark & \cmark & 633 \\
    
    \textcolor{darkgreen}{[D6]} - Huddl~\cite{paper20} & Football & \xmark & \xmark & \cmark & \cmark & 58 \\
    
    \textcolor{darkgreen}{[D7]} - Chinese Super League~\cite{paper37} & Football & \xmark & \xmark & \cmark & \cmark & 237 \\
    
    \textcolor{darkgreen}{[D8]} - German Bundesliga~\cite{paper42} & Football & \xmark & \xmark & \cmark & \cmark & 54 \\
    
    \textcolor{darkgreen}{[D9]} - NHL PBP~\cite{NHL_PBP} & Ice Hockey & \cmark & \xmark & \xmark & \cmark$^\ast$ & 9220 \\
    
    \textcolor{darkgreen}{[D10]} - SportLogiq\textsuperscript{\dag}~\cite{Sportlogiq} & Ice Hockey & \xmark & \xmark & \xmark & \cmark & 446 \\
    
    \textcolor{darkgreen}{[D11]} - NBA\textsuperscript{\dag}~\cite{NBA} & Basketball & \xmark & \xmark & \cmark & \cmark & 784  \\
    
    \textcolor{darkgreen}{[D12]} - World Tour~\cite{paper30} & Badminton & \cmark & \cmark & \cmark & \cmark & 21 \\
    
    \textcolor{darkgreen}{[D13]} - German League~\cite{paper32} & Handball & \xmark & \xmark & \cmark &  \xmark & 15 \\
    
    \textcolor{darkgreen}{[D14]} - NFL PBP~\cite{paper34} & American Football & \cmark & \xmark & \xmark & \cmark$^\ast$ & $\sim$256 \\
    
    \textcolor{darkgreen}{[D15]} - Table Tennis PBP~\cite{paper45} & Table Tennis & \xmark & \xmark & \xmark & \cmark$^{\ast\ast}$ & 152 \\

    \textcolor{darkgreen}{[D16]} - StatsPerform Rugby~\cite{cheradame2024expected} & Rugby & \xmark & \xmark & \xmark & \cmark & 1416 \\
    
    \toprule
\end{tabular}
}
\end{threeparttable}
\caption{Overview of datasets used in Action Valuation, detailing the sport, public availability, and data type. A more detailed version of the table is available in the supplementary material. $\dag$ indicates multiple dataset partitions used, $^\ast$ indicates event positional data precision loss (categorized into field zones), and $^{\ast\ast}$ indicates absence of positional data in events. \textsuperscript{1}Contains only the tracking of players within the camera's view, without the identification of those not in possession of the ball.}
\label{tab:datasets}
\vspace{-0.5cm}
\end{table}

These characteristics are often missing in publicly available datasets, as shown in Table~\ref{tab:datasets}, which provides an overview of commonly used datasets for AV, with a more detailed analysis in the supplementary material. Most public datasets only include ED, limiting methods to on-ball actions and lacking broader game context. Only two include OTD: one for badminton \textcolor{darkgreen}{[D12]}, where tracking data is extracted via a deep learning algorithm, and StatsBomb 360 \textcolor{darkgreen}{[D2]}, which tracks players within the camera view but only identifies the ball possessor. While this expands game context, it still limits the identification of players performing off-ball actions. StatsBomb 360 also has limited variability, focusing on specific teams rather than entire competitions. In contrast, private datasets, available through collaborations with companies and clubs, are generally more complete and enable more comprehensive modeling.



\vspace{0.2cm}
\noindent There is a clear gap between publicly and privately available datasets, leading to a similar disparity in AV research. This is driven by the competitive advantage AV applications can provide (see Section~\ref{sec:applications}), making teams and companies hesitant to share their data or methodologies. However, to advance research, there is an urgent need for publicly available, comprehensive datasets that enable fair comparisons and promote methodological progress.


\section{Methodologies}
\label{sec:methodologies}

In this section, we provide an overview of the various design characteristics of Action Valuation methods, summarized in Table~\ref{tab:methods}. Specifically, we examine the six distinct dimensions related to methodological choices: (T2.1) AV Framework, (T2.2) Architectural Modeling, (T2.3) Targeted Outcomes, (T2.4) Credit Assignment Horizon (CAH), (T2.5) Action Types, and (T2.6) Player-Aware Valuation.

\begin{table*}[t]
  \centering
  \rowcolors{3}{gray!25}{white}
  \resizebox{\textwidth}{!}{
  \begin{tabular}{lccccccccccc}
    \toprule

    Method & Year & Data & \multicolumn{2}{c}{Modeling} & Targeted Outcomes & CAH & \multicolumn{2}{c}{Action Types} & P-A & Evaluation & Applications \\
    \cline{4-5} \cline{8-9}
     & & & Spatial & Temporal & & & On-ball & Off-ball & &  \\
    \bottomrule
    \multicolumn{6}{l}{\textbf{Expectation-Based Framework}}& \textit{Future window} & & & & & \\
    \toprule
    \citet{paper23} & 2016 & \textcolor{darkgreen}{[D1]} & \multicolumn{2}{c}{SVM} & Shots & - & \cmark & \xmark & \xmark & FIT, RNK & PA \\
    
    \citet{paper5} - STARSS & 2017 & \textcolor{darkgreen}{[D1$^{\ast}$]} & \multicolumn{2}{c}{-} & Goals & - & \cmark & \xmark & \xmark & RNK, COR & PA \\
    
    \citet{paper20} & 2018 & \textcolor{darkgreen}{[D6]} & \multicolumn{2}{c}{Physics-based} & Goals & 1 act. & \cmark & \xmark & \xmark &  - & PA, AoG \\
    
    \citet{paper2} - VAEP & 2019 & \textcolor{darkgreen}{[D1$^{\ast}$]} & \multicolumn{2}{c}{CatBoost} & Goals & 10 act. & \cmark & \xmark & \xmark & FIT, RNK & PA, AoG\\
    
    \citet{paper33} & 2019 & \textcolor{darkgreen}{[D10]} & \multicolumn{2}{c}{Regression Tree} & GIM & - & \cmark & \xmark & \xmark & FIT, RNK & PA \\
    
    \citet{paper26} & 2019 & \textcolor{darkgreen}{[D11]} & - & LSTM & Points & 5 Sec. & \cmark & \xmark & \cmark & CAL & AoG \\
    
    \citet{paper34} & 2019 & \textcolor{darkgreen}{[D14]} & \multicolumn{2}{c}{Mult. Log. Reg.} & Points, Win & Sc. Pos. & - & - & \xmark & CAL, RNK & PA \\
    
    \citet{paper38} & 2020 & \textcolor{darkgreen}{-} & Graph Emb. & TF & Win & Game & \cmark & \xmark & \xmark & FIT & AoG \\
    
    \citet{paper36} & 2021 & \textcolor{darkgreen}{[D5]} & \multicolumn{2}{c}{CNN + DNN} & Goals & 15 Sec. & \cmark & \cmark & \xmark & CAL, FIT & AoG \\
    
    \citet{paper16} & 2021 & \textcolor{darkgreen}{[D1]} & \multicolumn{2}{c}{DNN} & xT & - & \cmark & \xmark & \xmark & FIT, RNK & PA \\
    
    \citet{paper32} & 2021 & \textcolor{darkgreen}{[D13]} & - & TF & Goals & 3 Sec. & - & - & \xmark & FIT, CAL & AoG, DMO \\
    
    \citet{paper22} & 2022 & \textcolor{darkgreen}{[D4]}  & \multicolumn{2}{c}{XGBoost} & Rec., EA & 5 Act. & \cmark & \xmark & \xmark & FIT, COR & PA, AoG \\
    
    \citet{paper37} & 2023 & \textcolor{darkgreen}{[D7]} & \multicolumn{2}{c}{LightGBM} & Velocity & - & \xmark & \cmark & \xmark & RNK, COR & PA \\ 
    
    \citet{paper6} - GNN-VSP & 2024 & \textcolor{darkgreen}{[D1$^{\ast}$]}  & \multicolumn{2}{c}{GNN} & Shots & - & - & - & \xmark & RNK & PA, AoG \\  

    \citet{cheradame2024expected} & 2024 & \textcolor{darkgreen}{[D16]} & \multicolumn{2}{c}{GAMM} & Points & 2 Pos. & \cmark & \xmark & \xmark & FIT, RNK & PA\\
    
    \bottomrule
    \multicolumn{6}{l}{\textbf{Markov Decision Process Framework}} & \textit{$\gamma$ [episode]} & & & & & \\
    \toprule
    \citet{xThreat} - xThreat & - & \textcolor{darkgreen}{-}  & \multicolumn{2}{c}{-} & Goals & 1 [5 Act.] & \cmark & \xmark & \xmark & RNK & PA \\
    
    \citet{paper45} & 2010 & \textcolor{darkgreen}{[D15]}  & \multicolumn{2}{c}{-} & Point & 1 [Point] & \cmark & \xmark & \xmark & - & AoG \\
    
    \citet{paper7_} & 2015 & \textcolor{darkgreen}{[D9$^{\ast}$]}  & \multicolumn{2}{c}{-} & Goals & 1 [Sc. Pos.] & \cmark & \xmark & \xmark & RNK, COR & PA, AoG \\
    
    \citet{paper25} & 2016 & \textcolor{darkgreen}{[D11]} & \multicolumn{2}{c}{-} & Points & 1 [Pos.] & \cmark & \xmark & \xmark & CAL, RNK & PA, DMO \\
    
    \citet{paper7} & 2017 & \textcolor{darkgreen}{[D10]}  & \multicolumn{2}{c}{-} & Goals, Win & 1 [14 Act.] & \cmark & \xmark & \xmark & COR & PA, AoG \\
    
    \citet{paper40} & 2021 & \textcolor{darkgreen}{[D1]}  & \multicolumn{2}{c}{-} & Goals & 1 [Pos.] & \cmark & \xmark & \xmark & FIT, CAL & PA, DMO \\
    
    \toprule
    \multicolumn{6}{l}{\textbf{Reinforcement Learning Framework}} & \textit{$\gamma$ [episode]} & & & & & \\
    \bottomrule
    \citet{paper8} - GIM & 2018 & \textcolor{darkgreen}{[D10]}  & - & LSTM & Goals & 1 [Sc. Pos.] & \cmark & \xmark & \xmark & RNK, COR & PA \\
    
    \citet{paper14} & 2020 & \textcolor{darkgreen}{[D1]}  & - & LSTM & Goals & 1 [Sc. Pos.] & \cmark & \xmark & \xmark & CAL, COR & PA, AoG \\
    
    \citet{paper10} & 2020 & \textcolor{darkgreen}{[D10]}  & \multicolumn{2}{c}{Linear layer} & Goals, IRL & 0.9 [Sc. Pos.] & \cmark & \xmark & \xmark & FIT, COR & PA \\
    
    \citet{paper42} & 2021 & \textcolor{darkgreen}{[D8]}  & \multicolumn{2}{c}{GRNN} & Enter dang. zone & - [Pos.] & \cmark & \cmark & \xmark & FIT, RNK & PA, AoG \\
    
    \citet{paper13} - RiGIM & 2022 & \textcolor{darkgreen}{[D10]} & CNN & LSTM & Goals & 1 [Sc. Pos.] & \cmark & \xmark & \xmark & CAL, COR & PA \\
    
    \citet{paper27} & 2022 & \textcolor{darkgreen}{[D1$^\ast$, D10]}  & - & LSTM & Points, Turn., FT\% & 0.8 [Pos.] & \cmark & \xmark & \xmark & RNK & PA, AoG, DMO \\
    
    \citet{paper30} & 2022 & \textcolor{darkgreen}{[D12]}  & - & LSTM & Point & 0.3 [Point] & \cmark & \xmark & \xmark & FIT, COR & AoG \\
    
    \citet{paper12} & 2023 & \textcolor{darkgreen}{[D4]}  & - & GRU & Goals, EPV, Actions & 1 [Pos.] & \cmark & \cmark & \xmark & FIT, COR & PA \\
    
    \citet{paper39} & 2024 & \textcolor{darkgreen}{[D3]}  & CNN & GRU & Phase-based outcome & 0.99 [Pos.] & \cmark & \xmark & \xmark & FIT & PA, AoG, DMO \\

    \toprule
\end{tabular}
}
\caption{Overview of Action Valuation methods categorized by (T2.1) AV Framework and sorted by year, detailing (T1) Data, (T2.2) Architectural Modeling, (T2.3) Targeted Outcomes, (T2.4) Credit Assignment Horizon (CAH), (T2.5) Action Types, (T2.6) Player-Aware consideration, (T3) Evaluation, and (T4) Applications. \textcolor{darkgreen}{$^\ast$}Indicates similar format data with unspecified details. Abbreviations: P-A-Player-Aware AV, TF—Transformer, Rec.—Ball Recovery, EA—Effective Attack, Turn.—Turnover, FT—Free Throw, Act.—Action, Sec.—Seconds, Pos.—Possession, Sc. Pos.—Scoring Possession. Evaluation and application abbreviations are in Sections~\ref{sec:evaluation} and~\ref{sec:applications}, respectively.}
\label{tab:methods}
\vspace{-0.5cm}
\end{table*}

\subsection{AV Framework}

The different frameworks used to estimate state and action values can be categorized into three main approaches.

\vspace{-0.4cm}
\paragraph{Expectation-Based (EB).} This approach assigns values to each state based on the expected number of desired outcomes occurring in a future time window, expressed as $V(S_t) = \mathbb{E}\left[N_{O_{\{t, t + \Delta t\}}} \mid S_t \right]$, where $N_{O_{\{t, t + \Delta t\}}}$ indicates the number of desired outcomes within the interval $(t, t + \Delta t)$, and $\Delta t$ represents the future window length. When considering only the occurrence or absence of a desired outcome ---i.e., $N_{O_{\{t, t + \Delta t\}}}$ is binary--- this simplifies to the probability of the event occurring, $V(S_t) = P(O_{\{t, t + \Delta t\}} \mid S_t)$, where $O_{\{t, t + \Delta t\}}$ is an indicator of the desired outcome occurring within the interval. To provide larger context when estimating these values, most methods incorporate information from a preceding time window $(t - \Delta t, t]$ rather than relying solely on static state features. Due to methodological similarities, we classify \cite{paper23}, \cite{paper5}, and \cite{paper6} under this framework, even though they estimate outcomes within an observed window or at the current state rather than in the future. As detailed in Section~\ref{sec:architectural}, most methods estimate these values via machine learning~\cite{paper23, paper33, paper34, paper22, paper37} or deep learning~\cite{paper26, paper38, paper36, paper16, paper32, paper6}, while some~\cite{paper5} avoid model training by using nearest-neighbor approaches to estimate probabilities. 

\vspace{-0.4cm}
\paragraph{Markov Decision Processes (MDP).} The problem described in Section~\ref{sec:task} aligns closely with MDPs~\cite{puterman1990markov}, defined by \textit{states} ($\mathcal{S}$), \textit{actions} ($\mathcal{A}$), \textit{rewards} ($R(s, a)$), and \textit{state transition probabilities} ($P(s' \mid s, a)$). In our case, these correspond to the game state, player actions, desired outcomes, and transition probabilities learned from the data. This section discusses methods assuming known MDP dynamics (via precomputed transition probabilities), in contrast to Reinforcement Learning approaches discussed later, which assume unknown dynamics and aim to learn them interactively. Given a known MDP, the value of a state or an action can be quantified as $V(S_t) = \mathbb{E}_{\pi}\left[\sum_{\tau=0}^{\infty}\gamma^{\tau}R(S_{t+\tau}, A_{t+\tau}^{\pi}) \mid S_t \right]$, and $V(A_t\mid S_t) = \mathbb{E}_{\pi}\left[\sum_{\tau=0}^{\infty}\gamma^{\tau}R(S_{t+\tau}, A_{t+\tau}^{\pi}) \mid S_t, A_t \right]$, respectively. Here, $\pi$ is the policy (the probability of selecting specific actions in given states), and $\gamma$ is the discount factor that weighs future rewards relative to immediate ones. These value can be computed via Dynamic Programming~\cite{xThreat, paper7_, paper7} using the Bellman equation. To estimate transition probabilities and policies from observed data frequencies, state and action spaces are often discretized, either with fixed field grids or through clustering algorithms that tailor grids for each action~\cite{paper7}. Alternatively, \cite{paper40} estimate these probabilities using a Bayesian approach. We also include \cite{paper25} in this category, even though it models the environment as a more general stochastic process rather than as an MDP.

\vspace{-0.4cm}
\paragraph{Reinforcement Learning (RL).} RL extends MDPs to cases where environment dynamics are unknown and must be learned interactively from data. In sports, where environment control is not feasible, most RL approaches use on-policy learning. Many are also value-based, estimating the value of a state, $V(S_t)$, or the value of taking an action at a state, $V(A_t\mid S_t)$. Alternatively, \cite{paper39} optimize the policy directly using Policy Gradient. Value-based methods often rely in Temporal Difference (TD) learning, which incrementally updates value estimates by combining observed rewards with predictions of future values. Some use the SARSA algorithm~\cite{paper8, paper14, paper30, paper12}, others the $\lambda$-return algorithm~\cite{paper42}, or even model a distribution over values via distributional TD learning~\cite{paper13}. Finally, \cite{paper27} combine value-based and policy-based methods using an actor-critic architecture, where the actor refines the policy and the critic, trained with TD learning, evaluates the current policy and provides feedback.



\vspace{0.2cm}
\noindent While early AV methods relied on MDPs for their simplicity, recent approaches primarily use EB and RL frameworks. EB methods typically use standard supervised learning, making them simpler and more interpretable than RL. In contrast, RL, though more complex, inherently captures sequential dependencies and generally handles long-term rewards more effectively. Multi-agent RL~\cite{zhang2021multi} can also help learn how to distribute value among concurrent player actions, while EB methods require predefined strategies. Thus, while EB is preferred for interpretability and ease of implementation, RL is better suited for AV due to its alignment with the task's sequential nature.

\subsection{Architectural Modeling}
\label{sec:architectural}
In this section, we compare the architectural designs of different approaches, focusing on whether they mainly model spatial information, temporal information, or both—either by simply aggregating data or using architectures designed to process spatial and temporal information simultaneously.

A common approach in expectation-based methods is to concatenate features from an observed time window and process them together using Machine Learning (ML)~\cite{paper23, paper2, paper33, paper34, paper22, paper37, cheradame2024expected} or Deep Learning (DL)~\cite{paper16} architectures. For instance, \cite{paper33} use a Regression Tree to distill knowledge from a RL approach~\cite{paper8}, to obtain a more interpretable model, while \cite{paper6} model spatial and temporal information together by aggregating data into a graph representing a dynamic pass network processed by a Graph Neural Network (GNN). Other methods~\cite{paper20, paper36} decompose the probability of a desired future outcome into multiple steps. \cite{paper36} incorporate spatiotemporal features to train Convolutional Neural Network (CNN)- and Deep Neural Network (DNN)-based models for predicting the next action and its expected value, while \cite{paper20} uses a physics-based model to compute state values through transition, control, and scoring probabilities. Other approaches focus on modeling temporal information within the preceding time window using architectures like Transformers~\cite{paper32} or LSTMs~\cite{paper26}. Additionally, \cite{paper38} explicitly model temporal information through Transformers while embedding spatial information from optical tracking data into a graph representation, which is iteratively updated.

In RL approaches, the sequential nature of the problem is typically addressed using architectures designed to model temporal information, with LSTMs~\cite{paper8, paper14, paper13, paper27, paper30} and GRUs~\cite{paper12, paper39} being the most common. Some methods~\cite{paper13,paper39} first process spatial information using CNNs before passing the features to a temporal module, while \cite{paper42} represent player positions as a graph and use a Graph Recurrent Neural Network (GRNN) to simultaneously model spatial and temporal information. Alternatively, \cite{paper10} take a simpler approach by training only linear layers in their Inverse RL setting.

\vspace{0.2cm}
\noindent Explicit spatial information modeling is specially relevant for OTD, containing detailed player position data. GNNs excel in this context, representing players as nodes and their relationships via edges. Geometric DL~\cite{wang2024tacticai} can also efficiently address field symmetries. For temporal sequences, architectures designed for sequential data—such as RNNs (e.g., GRUs or LSTMs) or Transformers—are more appropriate. When modeling both spatial and temporal data, further analysis is needed to determine whether to process spatial information first or handle both simultaneously.

\subsection{Targeted Outcomes}

As previously mentioned, AV values actions based on the occurrence of desirable outcomes. Alternatively, the task can be framed as avoiding undesirable actions, such as conceding goals, losing points, or losing possession~\cite{paper27}. Several methods jointly model both objectives, maximizing desirable outcomes while simultaneously avoiding undesirable ones~\cite{paper2, paper8, paper12, paper13, paper14, paper33, cheradame2024expected}.

An important consideration in AV is defining these desirable or undesirable outcomes, which can vary by sport and approach objectives. In games, the primary performance indicator is often the match outcome --winning or losing~\cite{paper34, paper38, paper7}. However, this supervision is often too sparse and overlooks the value of actions when the game's result is already settled. To address this, most approaches combine or replace it with events that directly influence the score, as they occur more frequently. These include goals~\cite{paper5, paper20, paper2, paper36, paper32, xThreat, paper7_, paper7, paper40, paper8, paper14, paper10, paper13, paper12, paper39} in sports like football, ice hockey, and handball, or points~\cite{paper26, paper34, paper45, paper27, paper30, cheradame2024expected} in sports such as basketball, badminton, American football, table tennis, and rugby.

In some sports, these events can still be infrequent, such as football with an average of three goals per game and ice hockey with six. This sparsity in desired outcomes makes it challenging to learn the value of actions comprehensively, often focusing on actions directly tied to scoring while neglecting earlier contributions. To mitigate this, some methods focus on more frequent events typically associated with positive outcomes, even if this is not always the case. Examples include shots~\cite{paper23, paper6}, recovering possession~\cite{paper22}, creating effective attacks~\cite{paper22}, or entering dangerous zones~\cite{paper42}. For instance, \cite{paper39} define desired outcomes based on play phases: moving the ball away from opponents during the transition phase, advancing the ball in the build-up phase, retaining possession in the established possession phase, and scoring a goal in the attacking phase. Other methods~\cite{paper12, paper16, paper33} attempt to distill information from existing valuation methods, using them to supervise their own approach. Some go further by treating all actions performed by professional players as positive examples, assuming these players as ``experts''. This is combined with goal supervision, either through Inverse Reinforcement Learning (IRL)~\cite{paper10} or by an additional temporal action classification loss~\cite{paper12}.

\vspace{0.2cm}
\noindent The most widely accepted desired outcomes in AV are those that directly impact game results, such as winning or scoring goals or points. However, due to reward sparsity, additional outcomes must be selected carefully. While generally considered positive, events like shots or entering dangerous zones may not always provide reliable supervision (e.g., a long-range shot with little chance of success or a pass into a dangerous zone with minimal teammate support). A promising direction is combining supervision from universally accepted desired outcomes with less obvious positive outcomes. Using all actions as positive supervision seems suboptimal, given the fast-paced nature of sports, where many in-game actions are not necessarily optimal. Additionally, considering both, desirable and undesirable outcomes is a valuable aspect of AV methods, enabling the consideration of both offensive and defensive valuation.

\subsection{Credit Assignment Horizon}

This section examines how far desired outcomes influence past state or action values, determining whether methods prioritize immediate or long-term rewards.

In EB approaches, the evaluated future window length determines how far ahead desired outcomes influence supervision. Most methods use a short look-ahead, either in time~\cite{paper26, paper36, paper32} --typically 3 to 15 seconds-- or in actions~\cite{paper2, paper22, paper20}, ranging from the next immediate action~\cite{paper20} to up to 10 actions. However, some consider entire scoring possessions (i.e., sequences that continue until a player or team scores)~\cite{paper34} or even full games~\cite{paper38}.

In MDP and RL approaches, the credit assignment horizon is determined by the discount factor, $\gamma \in (0,1)$, and the episode length into which games are divided, with higher $\gamma$ values emphasizing long-term rewards. Episodes often correspond to possessions (i.e., sequences of play where the same team retains control of the ball) or scoring possessions, restricting reward propagation within these bounds. MDP methods often set $\gamma=1$ but constrain reward propagation to single point~\cite{paper45}, possession~\cite{paper25, paper40}, or scoring possession~\cite{paper7_} episodes. Some~\cite{xThreat, paper7} limit the iterations when solving the Bellman equation to restrict the look-ahead to a fixed number of actions. Similarly, RL approaches typically use high $\gamma$ values with possession~\cite{paper12, paper39} or scoring possession~\cite{paper8, paper14, paper13} episodes, though a few prioritize short-term rewards with lower discount factors~\cite{paper10, paper27, paper30}.

\vspace{0.2cm}
\noindent There is no clear consensus on whether to prioritize short-term or long-term rewards. The choice is often subjective, depending on whether desired outcomes should influence only immediate actions or extend over a longer horizon. While there is clear interest in analyzing long-term rewards, emphasizing them can make it more challenging to accurately attribute outcomes to distant actions.




\subsection{Action Types}

An important distinction in this task is the type of actions being valuated. While on-ball actions are typically the most analyzed, off-ball actions, such as making runs or closing space, also have a crucial impact on the game and should be considered. Additionally, studies on on-ball actions often show a bias toward offensive plays, with defensive actions being underrepresented, as defenders primarily contribute through positioning and movement to disrupt the opponent's attack, rather than through ball control.

However, data availability is a key factor in assessing off-ball actions. Methods relying solely on event data, which typically capture only on-ball actions, miss the opportunity to evaluate off-ball movements. As a result, many approaches~\cite{paper23, paper2, paper33, paper38, paper16, xThreat, paper45, paper7_, paper7, paper40, paper8, paper14, paper10, paper13, cheradame2024expected} define the set of actions between states as only the on-ball action at that moment, $A_{St} = \{a_1\}$, with the time difference between consecutive states determined by the time between consecutive on-ball actions. This simplification attributes all value changes to the on-ball action, ignoring the impact of simultaneous off-ball actions. Similarly, some methods~\cite{paper26, paper22, paper25, paper27, paper30, paper39} using OTD still focus on on-ball actions, relying on tracking data only for additional state context. Other approaches, however, leverage tracking data to estimate off-ball action values. For instance, \cite{paper36} do not directly evaluate off-ball actions but assess player positioning by estimating how the game state value would change if a player received the ball at their location. \cite{paper37} focus on defensive off-ball actions, assuming defenders aim to move faster than expected when reacting to an attack. They compare actual player velocity with predicted velocity --representing the average defender’s speed-- to quantify defensive impact. Alternatively, \cite{paper42} train a trajectory prediction model and compute off-ball action value as the difference in state value between the performed and predicted trajectories. Lastly, \cite{paper12} integrate off-ball actions directly into state and action value learning, employing an independent multi-agent RL approach that distributes rewards across simultaneous on-ball and off-ball actions via TD learning.

\vspace{0.2cm}
\noindent The valuation of off-ball actions has been clearly understudied compared to on-ball actions, mainly due to data limitations and the complexity of assessing their impact. Future AV research should focus on valuing both on-ball and off-ball actions simultaneously. The main challenge lies in attributing state value changes to multiple simultaneous actions. While predefined rules can be used for value distribution, a more promising approach is to learn this assignment directly, as explored in multi-agent RL settings --whether with independent~\cite{paper12} or coordinated agents.


\subsection{Player-Aware Valuation}

A final consideration in AV is whether to include player attributes that may influence state or action value estimation. Most methods omit this, leading to models that generate values based on average players. Notably, \cite{paper26} is the only studied method that integrates player-specific information by learning player embeddings updated during training to account for individual identities when estimating values.

\vspace{0.2cm}
\noindent Incorporating player skills and characteristics into AV methods can be highly beneficial in certain applications (see Section~\ref{sec:applications}). For instance, optimizing players' decision-making can benefit from understanding the skills of the ball carrier, teammates, and opponents to determine the best action. However, for performance analysis, the role of player awareness is less clear. While recognizing opponent skill levels can help attribute higher value to actions against tougher rivals, most methods estimate value based on the difference between a player’s action and an average player’s baseline. Including player-specific data could alter state values, potentially undervaluing highly skilled players' actions. A possible approach is to account for all player skills except the one being valuated, though this complicates simultaneous action valuation. Clearly, player-aware valuation remains underexplored and requires further research.


\section{AV Evaluation}
\label{sec:evaluation}

Evaluating AV is challenging due to the inherently subjective nature of action value estimation, which results in a lack of ground-truth annotations and complicates the establishment of objective performance benchmarks. As a result, the methods assess their performance by evaluating their fit to the data or relying on subjective criteria (see Table~\ref{tab:methods} for an overview). Here we list the four most common criteria.

\vspace{-0.4cm}
\paragraph{Model Fit to Data (FIT):} Assessing how well the model fits the data~\cite{paper23, paper2, paper33, paper38, paper36, paper16, paper32, paper22, paper40, paper10, paper42, paper30, paper12, cheradame2024expected}, often by evaluating the loss function on unseen data to measure generalization. A key limitation is the lack of comparability across approaches, as fitness is typically measured using different metrics depending on the method. 

\vspace{-0.4cm}
\paragraph{Calibration Analysis (CAL):} Evaluating whether the generated state values meaningfully correlate with future desired outcomes by ensuring similar distributions between predicted values and observed outcomes~\cite{paper26, paper34, paper36, paper32, paper25, paper40, paper14, paper13}. While this is a fundamental step in validating the proper generation of state or action values, it does not enable comparisons across different methods. 

\vspace{-0.4cm}
\paragraph{Subjective Analysis of Player or Team rankings (RNK):}\footnote{Player and team scores can be derived from action values as described in Section~\ref{sec:applications}, enabling the creation of rankings.} Evaluating the rankings produced by the methods through subjective analysis, such as discussing players' contributions to top teams or highlighting relevant statistics alongside the rankings~\cite{paper5, paper2, paper33, paper34, paper16, paper37, paper6, xThreat, paper7_, paper25, paper8, paper42, paper27}. Additionally, some approaches justify player or team positions based on awards~\cite{paper5} or by comparing them to expert-curated top-player/team lists.

\vspace{-0.4cm}
\paragraph{Correlation with Standard Success Metrics (COR):} Using action values to compute player or team scores and evaluating their relationship with standard success metrics (e.g., goals, assists, points)~\cite{paper5, paper22, paper37, paper7_, paper7, paper8, paper14, paper10, paper13, paper30, paper12}. These metrics are measured within games used to train the valuation methods or future games, resembling a predictive capability analysis of the computed scores in relation to these metrics. While this approach provides a quantitative measure, it heavily depends on the predefined success metrics, which often overlook many aspects of player performance and tend to focus on offensive contribution.

\vspace{0.2cm}
\noindent While the evaluation methods discussed above provide valuable insights, none serve as a standardized evaluation framework. As noted, model fit depends on the method used, calibration analysis lacks direct performance assessment, subjective rankings are non-quantitative, and correlation with success metrics overlooks key aspects of player performance. Alternatively, \cite{overmeer2025revisiting} proposed an evaluation benchmark based on expert selection of the most valuable game state between two options, but still relying on expert subjectivity. This lack of a unified evaluation standard complicates comparisons between approaches. To advance AV research, efforts should focus on establishing a standardized evaluation protocol and benchmarking existing methodologies. One possible direction is expanding the correlation-based evaluation to assess not only the predictive capabilities of player or team scores towards standard success metrics but also their ability to forecast future game outcomes. The idea is that a method capable of accurately valuing actions should also produce reliable player or team performance estimates. If these scores are meaningful, they should exhibit stronger predictive power in anticipating game results compared to less effective valuation methods.

\section{Applications}
\label{sec:applications}

AV applications can be categorized into three main groups.

\vspace{-0.4cm}
\paragraph{Performance Analysis (PA).} 
One common application of AV is analyzing player and team performance. The performance of a player $P_i$ can be quantified as the cumulative value of his actions, represented as $S_{P_i} = \sum_{a \in A_{P_i}} V(a)$, where $A_{P_i}$ is the set of actions performed by $P_i$. Team performance can be calculated similarly. These scores can be computed for individual games or aggregated over multiple games to provide an indicator of average performance over time. In such cases, scores are typically normalized to represent values on a per-game basis. This application is particularly valuable for analyzing player performance across a season, as well as for identifying promising players, thereby aiding the scouting process. Its relevance is well-noticed in the literature, both at the player level~\cite{paper2, xThreat, paper5, paper6, paper7_, paper8, paper10, paper12, paper13, paper14, paper16, paper20, paper23, paper25, paper27, paper28, paper33, paper34, paper40, paper42} and the team level~\cite{paper7, paper20, paper22, paper27, paper37, paper39, cheradame2024expected}.

\vspace{-0.4cm}
\paragraph{Analysis of the Game (AoG).}  
AV is also highly valuable for gaining deeper insights from games. For instance, during or after a game, the distribution of action or state values can be analyzed to evaluate the dominance of each team or player throughout the match~\cite{paper20, paper22, paper26, paper27, paper30, paper36, paper39}. Moreover, it can be used to identify key actions that lead to substantial increases or decreases in value during a play~\cite{paper20, paper32, paper38, paper42}, or to assess the values of different types of actions in varying game situations~\cite{paper3, paper6, paper7_, paper7, paper14, paper26, paper36, paper45}. Additionally, some studies explore the trade-off between the quantity and quality (i.e., high value) of actions performed by different players~\cite{paper2}, or to characterize the playing styles of players based on the actions they perform and their associated values~\cite{paper2}.

\vspace{-0.4cm}
\paragraph{Decision-Making Optimization (DMO).}  
AV can also play a key role in optimizing the decision-making process of players. Specifically, some methods are designed not only to estimate the value of observed actions but also to evaluate alternative actions. For example, in basketball, Cervone et al.~\cite{paper25} compare the EPV (i.e., state value) when shooting with the EPV in alternative scenarios where the player passes the ball, allowing for the assessment of shot satisfaction. Similarly, other approaches~\cite{paper27, paper32, paper39, paper40}, across multiple sports, evaluate alternative scenarios given the current state, considering all the possible actions, and recommend the one that maximizes the value of the resulting state. This application can be highly beneficial for understanding optimal decision-making in specific situations and for teaching players how to apply these concepts in the fast-paced environment of an actual game.


\vspace{0.2cm}
\noindent As observed, AV applications span a wide range of areas in sports analytics, providing valuable insights for teams, players, and analysts. However, while AV methods are highly useful, their design should align closely with their intended applications. For instance, methods for PA or AoG may prioritize consistency in value estimation, whereas those for DMO should incorporate mechanisms to evaluate alternative actions and recommend the optimal ones. Therefore, future research should not only advance AV methodologies but also account for the specific needs and constraints of their target applications.







\section{Discussion}
\label{sec:discussion}

In this survey, we analyzed trends and advancements in Action Valuation, categorizing the approaches according to the proposed taxonomy. We identified critical gaps in two areas: data and evaluation, which hinder the objective benchmarking of methods under consistent conditions. Regarding methodologies, we examined the predominant frameworks, explored architectural designs, and then reviewed the commonly targeted outcomes and the trade-offs between short- and long-term rewards. We also discussed the analysis of on-ball and off-ball actions, and recent efforts to assign value to both simultaneously. Additionally, we explored the role of player-aware valuation. Finally, we reviewed the diverse applications of AV, emphasizing its importance in Sports Analytics, thus justifying the need for this survey to synthesize the current research on the topic.


Based on the topics discussed, the most urgent areas for advancing the field are data and evaluation. A comprehensive public dataset that includes both OTD and event data, providing full game context, is essential for enabling method comparison. Such a dataset would help bridge the gap between public and private research, greatly benefiting AV progress. This need is closely tied to the establishment of an objective evaluation framework to move beyond subjective assessments. As discussed in Section~\ref{sec:evaluation}, one potential method is measuring the predictive power of player or team scores for future game outcomes. In conclusion, a public dataset paired with an objective evaluation framework would facilitate method benchmarking, help identify key AV characteristics, and advance research in the field.

From a methodological perspective, further research is needed on the simultaneous valuation of both on-ball and off-ball actions, as only a few current approaches address this, with multi-agent RL being a potential solution. Additionally, studying the benefits of player-aware valuation, depending on the target application, is crucial, particularly in areas such as Decision-Making Optimization. Further exploration of methods to tackle the sparsity problem in desired outcomes in certain sports, as well as the trade-off between short- and long-term rewards, would also provide valuable insights for refining AV methods.

\noindent {\textbf{Acknowledgements.}} This work has been partially supported by the Spanish project PID2022-136436NB-I00 and by ICREA under the ICREA Academia programme.

{
    \small
    \bibliographystyle{ieeenat_fullname}
    \bibliography{main}
}

\newpage

\maketitlesupplementary
\appendix

\section{Extended data analysis}

In Table~\ref{tab:datasets_supp}, we extend the datasets table from the main paper by providing additional details on the number of games, number of distinct actions, and variability in terms of included competitions, teams, and players. As shown, StatsBomb\footnote{https://statsbomb.com/} \textcolor{darkgreen}{[D1]} with ED is the largest public football dataset, covering up to $3433$ games and $34$ different action types. Additionally, it offers large variability, spanning multiple competitions and seasons, and thus including a diverse range of teams and players. On the other hand, when extending the dataset to OTD data, StatsBomb 360 \textcolor{darkgreen}{[D2]} includes only a subset of these games, specifically 394. Furthermore, these games are often released in groups, meaning that for a given competition and season, only matches from a specific team are included. This limits the dataset's applicability. Among private football datasets, \textcolor{darkgreen}{[D3, D5, D7]} are among the most comprehensive, both in terms of the number of games and action types available.

In other sports, such as ice hockey, there is a publicly available dataset \textcolor{darkgreen}{[D9]} with high variability and a large number of games. However, it is limited to ED data, lacking information about the position of the player performing the action and containing only five different action types. More complete datasets are typically provided by SportLogiq\footnote{https://www.sportlogiq.com/}. For basketball, while the NBA\footnote{https://www.nba.com/} previously made OTD data publicly available, this is no longer the case, making it difficult to access datasets with high variability and a large number of games. For other sports, its datasets tend to be smaller, and commonly including less number of games.

\begin{table*}[t]
  \captionsetup{width=\textwidth}
  \centering
  \rowcolors{4}{gray!25}{white}
  \begin{threeparttable}
  \resizebox{\textwidth}{!}{
  \begin{tabular}{lcccccccc}
    \toprule
    Dataset & Sport & Public & \multicolumn{3}{c}{Data Type} & Games & Actions & Variability \\
    \cline{4-6}
    \multicolumn{3}{c}{} & VD & OTD & ED \\
    \bottomrule
    
    \textcolor{darkgreen}{[D1]} - StatsBomb\textsuperscript{\dag} & Football & \cmark & \xmark & \xmark & \cmark & 3433 & 34 & \blackcircle\blackcircle\blackcircle\blackcircle\blackcircle \\
    
    \textcolor{darkgreen}{[D2]} - StatsBomb 360  & Football & \cmark & \xmark & \cmark\tnote{1} & \cmark & 394 & 34 & \blackcircle\blackcircle\blackcircle\whitecircle\whitecircle \\
    
    \textcolor{darkgreen}{[D3]} - Belgian Pro League & Football & \xmark & \xmark & \cmark & \cmark & 430 & - & \blackcircle\blackcircle\blackcircle\blackcircle\whitecircle\\
    
    \textcolor{darkgreen}{[D4]} - Meiji J1 League & Football & \xmark & \xmark & \cmark & \cmark & 55 & - & \blackcircle\blackcircle\whitecircle\whitecircle\whitecircle \\
    
    \textcolor{darkgreen}{[D5]} - STATS LLC & Football & \xmark & \xmark & \cmark & \cmark & 633 & $3^+$ & \blackcircle\blackcircle\blackcircle\blackcircle\whitecircle \\
    
    \textcolor{darkgreen}{[D6]} - Huddl & Football & \xmark & \xmark & \cmark & \cmark & 58 & - & \blackcircle\blackcircle\blackcircle\whitecircle\whitecircle\\
    
    \textcolor{darkgreen}{[D7]} - Chinese Super League & Football & \xmark & \xmark & \cmark & \cmark & 237 & - & \blackcircle\blackcircle\blackcircle\blackcircle\whitecircle \\
    
    \textcolor{darkgreen}{[D8]} - German Bundesliga & Football & \xmark & \xmark & \cmark & \cmark & 54 & - & \blackcircle\blackcircle\blackcircle\whitecircle\whitecircle \\
    
    \textcolor{darkgreen}{[D9]} - NHL PBP & Ice Hockey & \cmark & \xmark & \xmark & \cmark$^\ast$ & 9220 & 5 & \blackcircle\blackcircle\blackcircle\blackcircle\whitecircle \\
    
    \textcolor{darkgreen}{[D10]} - SportLogiq\textsuperscript{\dag} & Ice Hockey & \xmark & \xmark & \xmark & \cmark & 446 & 43 & \blackcircle\blackcircle\blackcircle\blackcircle\whitecircle \\
    
    \textcolor{darkgreen}{[D11]} - NBA\textsuperscript{\dag} & Basketball & \xmark & \xmark & \cmark & \cmark & 784 & - & \blackcircle\blackcircle\blackcircle\blackcircle\whitecircle \\
    
    \textcolor{darkgreen}{[D12]} - Badminton World Tour & Badminton & \cmark & \cmark & \cmark & \cmark & 21 & 9 & \blackcircle\blackcircle\whitecircle\whitecircle\whitecircle \\
    
    \textcolor{darkgreen}{[D13]} - German Handball League & Handball & \xmark & \xmark & \cmark &  \xmark& 15 & 0 & \blackcircle\whitecircle\whitecircle\whitecircle\whitecircle \\
    
    \textcolor{darkgreen}{[D14]} - NFL PBP & American Football & \cmark & \xmark & \xmark & \cmark$^\ast$ & $\sim$256 & - & \blackcircle\blackcircle\blackcircle\blackcircle\whitecircle \\
    
    \textcolor{darkgreen}{[D15]} - Table Tennis PBP & Table Tennis & \xmark & \xmark & \xmark & \cmark$^{\ast\ast}$ & 152 & 6-11 & \blackcircle\blackcircle\blackcircle\blackcircle\whitecircle \\

    \textcolor{darkgreen}{[D16]} - StatsPerform Rugby & Rugby & \xmark & \xmark & \xmark & \cmark & 1416  & - & \blackcircle\blackcircle\blackcircle\blackcircle\whitecircle \\
    
    \toprule
\end{tabular}
}
\end{threeparttable}
\caption{Overview of datasets used in Action Valuation, detailing the sport, public availability, data type, number of games, number of action classes, and variability in terms of included competitions, teams, and players. $\dag$ indicates multiple dataset partitions used, $^\ast$ indicates event positional data precision loss (categorized into field zones), and $^{\ast\ast}$ indicates absence of positional data in events. \textsuperscript{1}Contains only the tracking of players within the camera's view, without the identification of those not in possession of the ball. \textsuperscript{+}Represents the number of action classes used from a larger, unspecified set.}

\label{tab:datasets_supp}
\vspace{-0.5cm}
\end{table*}

\end{document}